\pdfoutput=1

\documentclass[a4paper,fleqn]{cas-sc}

\usepackage[numbers]{natbib}
\usepackage{lscape} 
\usepackage{amssymb}
\usepackage{amsmath}
\usepackage{algorithm}
\usepackage{algorithmic}
\usepackage{booktabs}
\usepackage{threeparttable}
\usepackage{multirow}
\usepackage{makecell}
\usepackage{bm}
\usepackage{hyperref}
\usepackage{makecell}
\usepackage{bm}
\usepackage{hyperref}

\usepackage{soul, color, xcolor}

\soulregister\cite7
\soulregister\ref7 
\soulregister\citep7 
\soulregister\citet7 
\soulregister\pageref7 
\soulregister\autoref7 

\makeatletter
\def\tagform@#1{\maketag@@@{\ignorespaces#1\unskip\@@italiccorr}}
\let\orgtheequation\theequation
\def\theequation{(\orgtheequation)}
\makeatother
\let\orgautoref\autoref

\renewcommand{\autoref}[1]{\def\equationautorefname{Eq.}\orgautoref{#1}}
\def\tsc#1{\csdef{#1}{\textsc{\lowercase{#1}}\xspace}}
\tsc{WGM}
\tsc{QE}

\begin{document}
\let\WriteBookmarks\relax
\let\printorcid\relax
\def\floatpagepagefraction{1}
\def\textpagefraction{.001}

\shorttitle{}    

\shortauthors{Kuang et al.}  

\title [mode = title]{Citywide Electric Vehicle Charging Demand Prediction Approach Considering Urban Region and Dynamic Influences}

\author[label1,label2]{Haoxuan Kuang}

\author[label1,label2]{Kunxiang Deng}

\author[label1,label2]{Linlin You}

\author[label1,label2]{Jun Li\corref{cor1}}
\ead{stslijun@mail.sysu.edu.cn}

\affiliation[label1]{organization={School of Intelligent Systems Engineering, Sun Yat-sen University},
	addressline={No.66, Gongchang Road, Guangming District}, 
	city={Shenzhen},
	postcode={518107}, 
	state={Guangdong Province},
	country={China}}

\affiliation[label2]{organization={Guangdong Provincial Key Laboratory of Intelligent Transportation System, School of Intelligent Systems Engineering, Sun Yat-Sen University},
	city={Guangzhou},
	postcode={510275}, 
	country={China}}

\cortext[cor1]{Corresponding author}

\begin{abstract}
Electric vehicle charging demand prediction is important for vacant charging pile recommendation and charging infrastructure planning, thus facilitating vehicle electrification and green energy development. The performance of previous spatio-temporal studies is still far from satisfactory nowadays because urban region attributes and multivariate temporal influences are not adequately taken into account. To tackle these issues, we propose a learning approach for citywide electric vehicle charging demand prediction, named CityEVCP. To learn non-pairwise relationships in urban areas, we cluster service areas by the types and numbers of points of interest in the areas and develop attentive hypergraph networks accordingly. Graph attention mechanisms are employed for information propagation between neighboring areas. Additionally, we propose a variable selection network to adaptively learn dynamic auxiliary information and improve the Transformer encoder utilizing gated mechanisms for fluctuating charging time-series data. Experiments on a citywide electric vehicle charging dataset demonstrate the performances of our proposed approach compared with a broad range of competing baselines. Furthermore, we demonstrate the impact of dynamic influences on prediction results in different areas of the city and the effectiveness of our area clustering method.
\end{abstract}


%


\maketitle

\section{Introduction}
As the world accelerates its transition to low-carbon and environmental friendly mobility, electric vehicles (EVs) will play an important role in the future automotive market \citep{powell2022charging}. The International Energy Agency points out that at present, the global share of electric vehicles in new car sales is about 20\%, and this proportion will rise to about 50\% by 2030, while China has already reached such a level in 2024 \citep{IEA2024}. However, the lack of available and reasonably located charging facilities is a major constraint to the development of electric vehicles \citep{you2024unraveling}. Providing unlimited charging infrastructure for EV users is impossible in terms of investment costs and energy deployment \citep{manriquez2020impact}. The increasing development of smartphones, in-vehicle devices and information distribution systems makes governments and mangers glad to turn to accurate EV charging demand prediction for help. By predicting the services demand  in the coming future, administrators can provide users with suggestive information, such as timing and route planning, to avoid customer losses. For governments, accurate charging demand prediction, especially at the citywide level, is a groundwork for a variety of smart strategies such as charging infrastructure planning and smart power supply \citep{ ding2020optimal, wu2019demand}.

Researchers have made many efforts to enable accurate citywide electric vehicle charging demand prediction. Deep learning research provides powerful methods for extracting complex patterns in data variations nowadays. Many studies consider that EV charging demand in future periods is related to historical demand with spatial spillover effects \cite{kuang2024unravelling}. Therefore, lots of exquisite models introduce Recurrent neural networks to extract temporal features \citep{wang2023short} and Graph neural networks to extract spatial features \citep{kipf2016semi}, respectively. However, the ambiguity in the weight assignment of convolutional and recurrent operations limits their capabilities. In the latest researches, time-series attention networks (e.g., Transformer) \citep{vaswani2017attention} and graph attention networks \citep{velivckovic2018graph} that are able to better determine the weights on the network have become popular tools for improving the spatio-temporal prediction accuracy. However, previous studies have lacked sufficient consideration of the characteristics of charging demand that are different from other spatio-temporal data \citep{arias2016electric}. They do not consider that some short periods of fast charging behavior may cause fluctuations in the historical data, which results in not all hidden states can provide enough useful information for prediction. Sparse and rational allocation of attention for accurate charging demand prediction in attention networks remains to be explored.

\begin{figure}[htbp]
	\centering
	\includegraphics[width=5in]{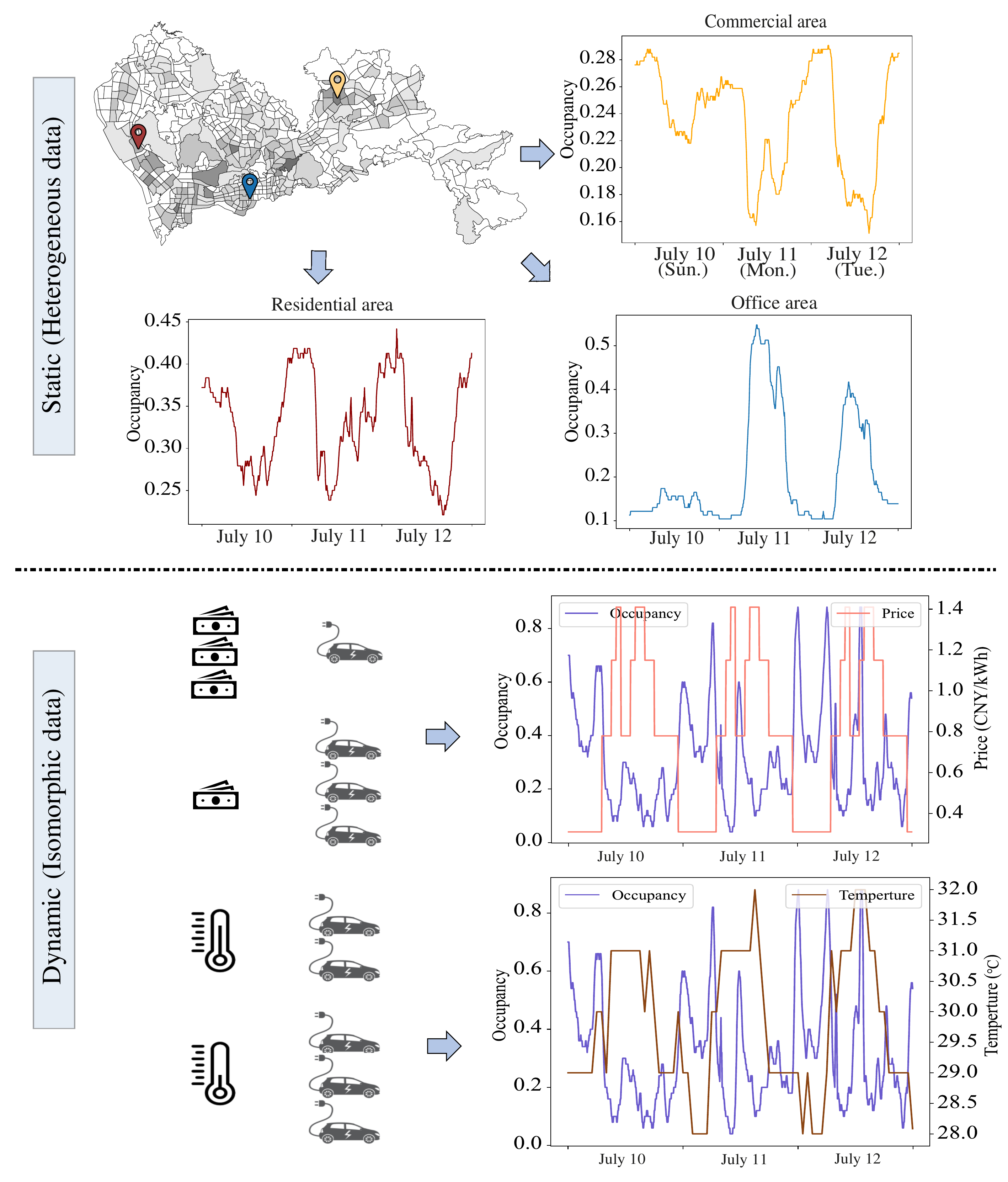}
	\caption{The impact of different factors on charging demand.}
	\label{dynamic}
\end{figure}

Furthermore, EV charging behavior is affected by several influencing factors, as shown in \autoref{dynamic}. With the application of information collection systems and large-scale data processing centers, various charging-related factors can be effectively centralized and analyzed. Considering adequate factors in urban EV charging demand prediction has become an urgent and promising topic. But there is still a lack of prediction studies that can effectively consider multiple influences at the same time. 

Charging-related factors can be categorized into two main groups, namely static ones and dynamic ones. Static influences include neighborhood charging services and land use around the charging stations. These influences are heterogeneous to the time-series charging demands and are usually not directly acceptable as inputs to the network. For example, improvements of the road network and the smart application on mobile devices make it possible for demand in neighboring areas to influence each other. Previous studies employ graph neural networks to fuse information from adjacent nodes and achieve good results \citep{xie2023spatio,zhang2024multimodal}. Moreover, some areas have similar land uses and residents may have typical and similar charging habits within those areas \citep{tahir2024sustainable}. For example, charging demand in commercial areas is relatively high on weekends and after work on weekdays, while it's the opposite in office areas. Clustering similar nodes and then training the network in separate groups is a common method used in previous studies \citep{wang2023predicting}. However, it is still a challenge to accurately cluster nodes with similar properties. And separating the network into multiple groups for training makes it difficult to share information between groups and consumes more computational resources and storage space.

Dynamic influences and demand data are usually isomorphic, both characterized by changes over time. These dynamics include charging price, charging environment, and other variables that affect charging demand in real time. For example, price influences the charging choices of price-sensitive users, and temperature largely affects the power consumption and charging efficiency of electric vehicles \citep{kuang2024unravelling}. The relationship between external information and charging demands is often complex and varied in temporal and spatial dimensions. Therefore we need to accurately capture the nonlinearities involved. Previous research mainly used convolution to fuse dynamic external data. Limited by the ambiguity, the convolution methods only bring a certain but weak improvement in prediction accuracy \citep{kuang2024physics, yang2019deeplearning}. It is still a challenge to determine the different roles of external auxiliary information in each urban area. A more precise method for determining the citywide significance of dynamic influences is needed.

To address the above challenges, we propose a novel approach for citywide EV charging demand prediction, named CityEVCP, that can effectively consider urban region and dynamic influences simultaneously. It enables 1) classifying areas based on an adaptive clustering method and simultaneously learning similar patterns from areas with similar properties through attentive hypergraph; 2) information propagation between neighboring areas through graph attention; and 3) fusing time-series external variables for city's each areas through variable selection network and efficiently capturing temporal features through an improved gated Transformer encoder. We demonstrate the state-of-the-art performance of the proposed CityEVCP method through a dataset collected from Shenzhen, a first-tier city in China. We also demonstrate the contribution of multiple  influences to accurate EV charging demand prediction through the case study and show that the proposed model can correctly perceive charging-related data patterns.

The main contributions of this paper are as follows.

\begin{itemize}
	\item We propose a novel deep learning framework for citywide electric vehicle charging prediction that can effectively consider charging-related dynamic and static influences. We incorporate an area attributes fusion module, an adjacency graph fusion module, and a temporal feature fusion module to achieve accurate prediction.

	\item We propose a scheme for considering urban properties around charging service facilities. We view the points of interest (POIs) in an area as a document and propose an adaptive clustering method based on the importance of POIs. Attentive hypergraph mechanism is used to learn common features of charging behavior in grouped urban areas.
	
	\item Considering the different role of dynamics on areas, we utilize gated residual networks (GRNs) to propose a nonlinear variable selection network that adaptively determines the importance of multiple dynamic influences in each urban area. Furthermore, to flexibly adapt to fluctuations in charging demand, we improve the Transformer encoder through GRNs and Gumbel-Softmax, which determines sparse attention for the timestamps and avoids fuzzy perception.

\end{itemize}

The remaining article is structured as follows. \autoref{work} reviews the relevant literature. \autoref{methodology} describes our proposed method, CityEVCP, in detail. The method is validated in \autoref{experiments} with a representative citywide EV charging demand dataset. \autoref{conclusion} draws conclusions and future directions.

\section{Literature review}
\label{work}
Citywide EV charging demand prediction can be considered as a spatio-temporal regression problem, which plays an important role in urban planning, charging guidance, and grid stabilization. Early studies mainly employed statistical methods such as Linear regression, Autoregressive Integrated Moving Average (ARIMA)\citep{williams1998urban} and Least absolute shrinkage and selection operator (Lasso)\citep{tibshirani1996regression}. Several studies have recognized that EV charging demand is associated with multi-source factors (e.g., price) and attempted to model the relationship using linear models \citep{zhu2018meta}. These models are simple and computationally fast, and make important attempts at modeling time-series EV charging demand and its correlates, but are limited by their linear assumptions, and the predictions are not precise enough. The development of machine learning has quickly made it a popular method for data regression. Various methods such as Random Forest and Support Vector Regression have been used for electric vehicle charging demand prediction and achieved good results \citep{breiman2001random, ge2020data}. However, these simple machine learning models still on one hand lack the consideration of complex nonlinear multivariate relationships, and on the other hand it is difficult to incorporate heterogeneous data (e.g., land attributes and spatial relationships) into the model.

Neural networks and deep learning can not only accurately model temporal relationships, but also learn multivariate and multi-node features, which become a research hotspot for citywide electric vehicle charging demand prediction. Recurrent neural networks (including Long Short Term Memory (LSTM)\citep{hochreiter1997long, ma2022multistep} and Gated Recurrent Unit (GRU)\citep{chung2014empirical}) make an important contribution to the accuracy of regression. A study proposes a new hybrid LSTM neural network that contains both historical charging state sequences and time-related features, achieving excellent time-series EV charging demand prediction \citep{wang2023short}. Graph Convolutional Neural Network (GCN) is an application of Convolutional Neural Networks (CNN) on graphs, which enables information from neighbors to be learned \citep{kipf2016semi}. The combination of graph neural networks and recurrent neural networks provides the model with powerful spatio-temporal modeling capabilities \citep{xie2023spatio}. A study combining GCN and GRU achieves excellent results on traffic prediction tasks \citep{zhao2019t}. A study improved the spatio-temporal learning capability of the model by replacing the linear operation in GRU with graph convolution \citep{zhang2023tmfo}. Attention mechanism enables the weights in the model to be determined rationally and accurately. Transformer can not only determine the importance of temporal feature, but also parallelize the hidden state processing to achieve a faster computation speed than recurrent neural networks \citep{vaswani2017attention, lim2021temporal}. Graph Attention Network (GAT) is a combination of GCN and attention mechanism that allow weights to be determined on graphs \citep{velivckovic2018graph}. An air traffic density prediction study combines GAT and LSTM to make accurate and robust predictions that outperform baseline models \citep{xu2023air}. However, previous research has focused mainly on how to skillfully combine existing tools, while ignoring data characteristics in specialized domains. In the task of EV charging demand prediction, few studies have specifically optimized the network structure in response to the short-term fluctuations in charging demand data.

Apart from EV charging demand data, more and more studies are beginning to focus on other influencing factors for assisting in accurate prediction \citep{zhang2021periodic, lu2024hyper, lu2024ev}. One of the major influencing factors is the price, which affects the users' choices. Incorporating price information into the model can dynamically capture user behavior and reduce prediction lag. A deep learning charging demand prediction study utilizes attention mechanisms and bi-directional recurrent neural networks to take into account the effect of price and achieve accurate predictions \citep{hu2024fractional}. Previous studies use mixed pseudo-sample meta-learning and physics-informed neural networks to model the relationship between charging demand and price, respectively, but these studies still lack consideration of other influencing factors \citep{qu2023physics, kuang2024physics}. Day and night show the same cyclical variation as temperature (i.e., high temperatures during the day and low temperatures at night), and EV power consumption is closely related to the environment (e.g. EVs consume more electricity when temperatures are low). In previous studies, environment variables are used to assist in predicting traffic behavior such as parking \citep{zhang2021periodic}, but there is still few charging demand prediction studies that take environment patterns into account. A study fine-tunes a large language model to enable few-shot predictions of citywide EV charging demand using socio-economic and environmental knowledge pre-trained by the large language model \citep{qu2024chatev}.

Land use information (e.g., points of interest) can help determine the amenities surrounding a charging station. A demand prediction study first learns spatio-temporal information by using GCN and GRU, then categorizes EV charging stations based on points of interest, and handles each class of charging stations separately using fully connected networks \citep{wang2023predicting}. However, handling grouped nodes separately is not efficient and hinders some of the information dissemination. Accurate methods for clustering and fusing intragroup information still remains to be explored. A recent multi-view joint graph learning study applies text sequence information retrieval method to identify regions' intrinsic key attributes and achieves outstanding results \citep{zhang2021multi}. Furthermore, existing graph neural network frameworks are deployed based on simple graph structures for one-by-one paired nodes, which limits their application when dealing with complex associations of grouped data. Recently hypergraph-based methods have been proposed to solve this problem \citep{feng2019hypergraph}. A general high-order multi-modal/multi-type modeling framework called HGNN+ is proposed to simultaneously handle multi-hop adjacencies, attribute groupings, and other hyperedge relations under a single hypergraph-based framework \citep{gao2022hgnn+}. Hypergraph convolutional neural networks that allow information to propagate over hypergraphs have a wide range of applications in both regression and classification tasks, such as multi-task traffic prediction \citep{wang2022multitask}, metro passenger flow prediction \citep{wang2021metro}, POI recommendation \citep{zeng2025global} and student academic performance classification \citep{li2022multi}. Moreover, several studies utilize attention mechanisms to improve hypergraph convolutional networks to select important attributes \citep{yang2022co,bai2021hypergraph}. Research on the application of these latest solutions to accurate citywide EV demand prediction remains to be explored.

In summary, deep learning enables models with powerful feature extraction capabilities, thus researchers are working on building exquisite and efficient deep learning models for accurate EV charging demand prediction. Recent researches are turning to advanced attention mechanisms to better determine the weights of the network. A growing number of studies consider multi-source factors related to EV charging demand. However, models that can comprehensively and effectively account for both urban region and dynamic external influences are still lacking. We still need to explore refined approaches to learn complex spatio-temporal features for accurate citywide EV charging demand prediction.

\section{Methodology}
\label{methodology}
\begin{figure}[htbp]
	\centering
	\includegraphics[width=6.5in]{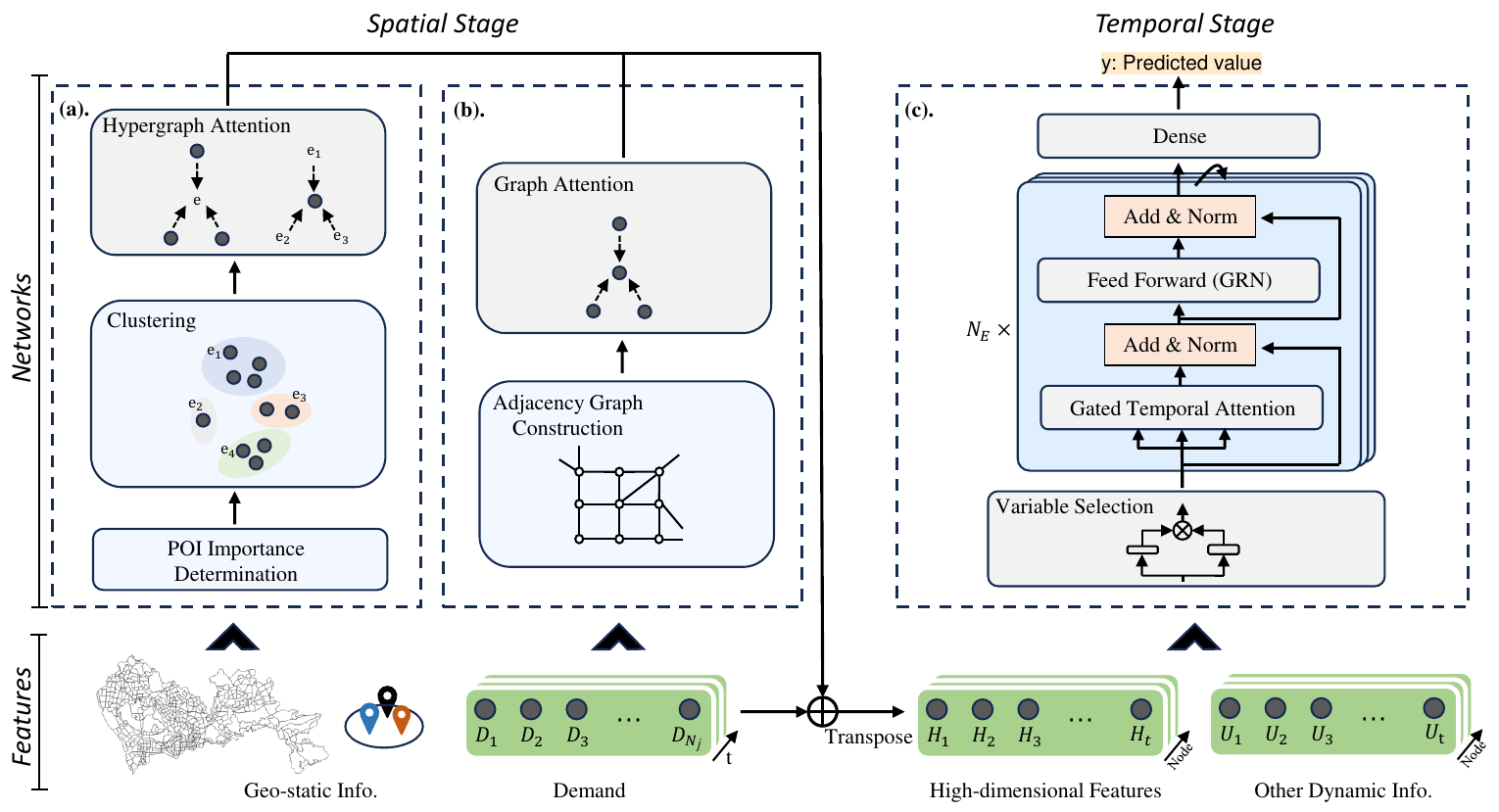}
	\caption{The model structure of the proposed approach, which consists of (a). Area attributes fusion module, (b). Adjacency graph fusion module, and (c). Temporal feature fusion module.}
	\label{overview}
\end{figure}

We divide a city into sub-areas based on travel community for the convenience of urban studies. Electric vehicle charging demand is considered the same as the number of EV charging piles occupied in a given sub-area or travel community. \autoref{overview} illustrates the structure of CityEVCP, which consists of three modules, namely a) an area attributes fusion module, in which areas with similar attributes are clustered according to POI importance, and grouped features are learned by using attentive hypergraph neural network; b) an adjacency graph fusion module, which allows for learning adjacency features through graph attention mechanism; and c) a temporal feature fusion module, which enables multi-source temporal variable selection and temporal feature learning.

\subsection{Area Attributes Fusion Module}
\label{AAFM}
Similar areas usually have similar traffic characteristics, e.g., residential locations usually have relatively high demand for charging at night. The type and number of POIs can be a good indicator of the area's utilization attributes, but it is not in the same form as demand data and cannot be used directly as an input to the model. We view POIs in an area as a document, as shown in \autoref{document}.

\begin{equation}
	\label{document}
	<\rm{POI}^j_1,\  \rm{POI}^j_2,\  \rm{POI}^j_3,......,\  \rm{POI}^j_n>
\end{equation}
where $\rm{POI}^a$ stands for POI terms in area $j$ and $n$ denotes the number of the POI terms.

Then, we count the number of POIs in each category. We treat a POI category as a word in documents. Information retrieval and mining technique is employed to measure the importance of POI categories for an area. 
In order to focus on both high-frequency occurrences and significantly differentiated POI categories, Term Frequency-Inverse Document Frequency (TF-IDF)\citep{salton1971smart} is chosen, where a high TF-IDF value is expected if the frequency of a POI type is high in a specific area and the frequency of that POI is low in the entire city. The calculation of TF-IDF is shown in \autoref{TF-IDF}.

\begin{equation}
	\label{TF-IDF}
	u_{i,j} = \frac{f_{i,j}}{\sum_{k}f_{k,j} }  \times   \log_{}{\frac{N_j }{1+\left | j:i\in j \right | } } 
\end{equation}
where $u_{i,j}$ denotes the Term Frequency-Inverse Document Frequency of POI category $i$ in area $j$, $f_{i,j}$ denotes the number of times POI category $i$ appears in area $j$, $N_j$ denotes the number of areas, and $\left | j:i\in j \right | $ denotes the number of areas containing POI category $i$.

From the above we obtain the sequence $U_{j}=\{u_{1,j},u_{2,j},...\}$ of the importance of each POI category for each area $j$. Then, we employ K-means clustering \citep{macqueen1967some} to divide the areas into a specific number of groups (denoted as $C$) based on TF-IDF sequences $U$.

In order to fuse features within groups on a single network, hypergraph attention mechanism is introduced. We consider all areas of the entire city as a hypergraph with incidence matrix $\bm{M}\in \left \{ 0,1 \right \} ^{\left | \nu  \right | \times \left | \varepsilon   \right |}$, where $\nu$ denotes the vertex set and $ \varepsilon$ denotes the hyperedge set. Besides, we consider each area group as a hyperedge $\bm{e}_i$, which belongs to $ \varepsilon$. Hyperedge $\bm{e}_i$ connects specific areas ${\left \{   \bm{n}_j \mid n_j\in e_i\right \} }$. Attentive hypergraph neural network is an improvement of hypergraph convolution network, which is divided into two steps. 

The first step is to propagate the information from the nodes within the cluster to the hyperedge, where we consider a entity-free hyperedge as a center node. The second step enables the node to obtain auxiliary information from the hyperedges connected to it. The calculation process is shown in \autoref{hypergraph}.

\begin{equation}
	\label{hypergraph}
	\begin{aligned}
		\bm{e}^{(1)}_i = {\rm{att}}(\bm{e}_i,\{\bm{n}_j\mid n_j\in e_i\})	\\
		\bm{n}^{(1)}_j = {\rm{att}}(\bm{n}_j,\{\bm{e}^{(1)}_i\mid n_j\in e_i\})
	\end{aligned}
\end{equation}
where $\bm{e}_i^{(1)} \in \mathbb{R} ^{1 \times \tau }$, $\bm{n}_j^{(1)} \in \mathbb{R} ^{1 \times \tau }$ denote high-dimensional features of hyperedges and nodes after calculation, respectively.

We take the first step as an example, showing how the update embedding $\bm{e}^{(1)}_i$ of a hyperedge is aggregated from itself and its connected nodes. The computational process is essentially the same as the graph attention mechanism, where features are concatenated and then feature-aware soft attention mechanism is used to determine the importance between nodes. The specific calculation process is shown in \autoref{softatten}.

\begin{equation}
	\label{softatten}
	\begin{aligned}
		\omega _{ij} ={\rm LeakyReLU}(\bm{W}_\omega(\bm{e}_{i} \parallel \bm{n}_{j}))\\
		{A_{ij}} ={\rm Softmax}_{j} \left ( \omega _{ij} \right ) =\frac{{\rm exp}(\omega _{ij} )}{ {\textstyle \sum_{k\in \mathcal{N}_{i} }^{}{\rm exp}(\omega _{ik})}}\\
		\bm{e}^{(1)}_i ={\rm LeakyReLU}( {\textstyle \sum_{j\in \mathcal{N}_{i} }^{}} {A_{ij}} \bm{n}_{j})
	\end{aligned}
\end{equation}
where $\omega _{ij}$ and ${A_{ij}}$ are the relationship variable and attention value of hyperedge $\bm{e}_i$ and node $\bm{n}_j$, respectively,  LeakyReLU is a nonlinear activation function, $\bm{W}_\omega$ is a coefficient matrix, "$\parallel$" means vector concat, and $\mathcal{N}_{i}={\left \{   \bm{n}_j \mid n_j\in e_i\right \} }$ stands for the set of i-th hyperedge's nodes.

Through attentive hypergraph neural network, we obtain high-dimensional features $\bm{H}_a$ that fuse similar land use attribute areas patterns.

\subsection{Adjacency Graph Fusion Module}
Well-established urban road networks and information dissemination platforms allow charging demand to spill over from one area to another, so that the demand in neighboring areas is closely correlated. We employ graph attention mechanism (GAT) to model the spatial feature of urban EV charging demand. The graph attention mechanism is an improvement of graph convolution, which enables weights on the graph to be determined and information propagation to be focused. First, we define the structure of an undirected graph $\bm{G}$. We define two areas to be connected when the two areas have at least one common edge, as shown in \autoref{spatially connected}.

\begin{equation}
	\label{spatially connected}
	a_{ij}^{G} = \begin{cases}
		1, &\text{if area } i \text{ and area } j \text{ are spatially neighboring}\\
		0 , &\text{otherwise}
	\end{cases}
\end{equation}
where $a_{ij}^{G} $ stands for the adjacent relationship of area $i$ and area $j$.

The calculation process of GAT is shown in \autoref{GAT_all}.

\begin{equation}
	\label{GAT_all}
	\bm{n}^{(1)}_i = {\rm{att}}(\bm{n}_i,\{\bm{n}_j\mid n_j\sim  n_i\})
\end{equation}
where $\bm{n}^{(1)}_i$ is the update embedding feature of node $i$, and $n_j\sim  n_i$ stands for node $i$ and node $j$ is connected.

The specific calculation steps are shown in \autoref{GAT}, which is similar to \autoref{softatten}. First features are concatenated, then soft attention between nodes is calculated, and finally the information is enabled to propagate on the graph based on the attention weights.

\begin{equation}
	\label{GAT}
	\begin{aligned}
		\omega^G _{ij} ={\rm LeakyReLU}(\bm{W}^G_\omega(\bm{n}_{i} \parallel \bm{n}_{j}))\\
		{A^G_{ij}} ={\rm Softmax}_{j} \left ( \omega _{ij} \right ) =\frac{{\rm exp}(\omega _{ij} )}{ {\textstyle \sum_{k\in \mathcal{N}_{i} }^{}{\rm exp}(\omega _{ik})}}\\
		\bm{n}^{(1)}_i ={\rm LeakyReLU}( {\textstyle \sum_{j\in \mathcal{N}_{i} }^{}} {A_{ij}} \bm{n}_{j})
	\end{aligned}
\end{equation}
where $\omega^G_{ij}$ and ${A^G_{ij}}$ are the relationship variable and attention value of node $\bm{n}_i$ and $\bm{n}_j$, respectively, and $\bm{W}^G_\omega$ is a coefficient matrix.

It is worth noting that using GAT to model spatial features here is actually a simplification of spatial hypergraph attention and achieves the same effect as it. When we consider a pair of nodes connected to each other as a hyperedge, the spatial feature can be treated as a hypergraph, but this brings unnecessary computation.

Through graph attention mechanism, we obtain high-dimensional features $\bm{H}_b$ that fuse adjacent areas patterns.

$\bm{H}_a$, $\bm{H}_b$ and sliced demand data $\bm{D}_s$ are added together to obtain high dimensional features $\bm{H} \in \mathbb{R} ^{N_{A}\times \tau}$, as shown in \autoref{high dimensional}. Note that the temporal information in them is preserved.

\begin{equation}
	\label{high dimensional}
	\bm{H} = {\rm{Add \& Norm}}(\bm{H}_a, \bm{H}_b, \bm{D}_s)
\end{equation}
where ${\rm{Add \& Norm}}$ stands for adding up the hidden states and layer normalization \citep{ba2016layer}.

\subsection{Temporal Feature Fusion Module}
To accurately predict the EV charging demand, it is worthwhile to take into account external information (e.g., price, temperature) that can influence the user's choice of charging service. Considering auxiliary information in the model is expected to effectively capture dynamic changes and reduce lags in prediction. The data structure of some external information is the same as the time-series demand data. In this subsection, we propose a temporal feature fusion scheme that is appropriate for urban EV charging demand prediction.

\begin{figure}[htbp]
	\centering
	\includegraphics[width=5in]{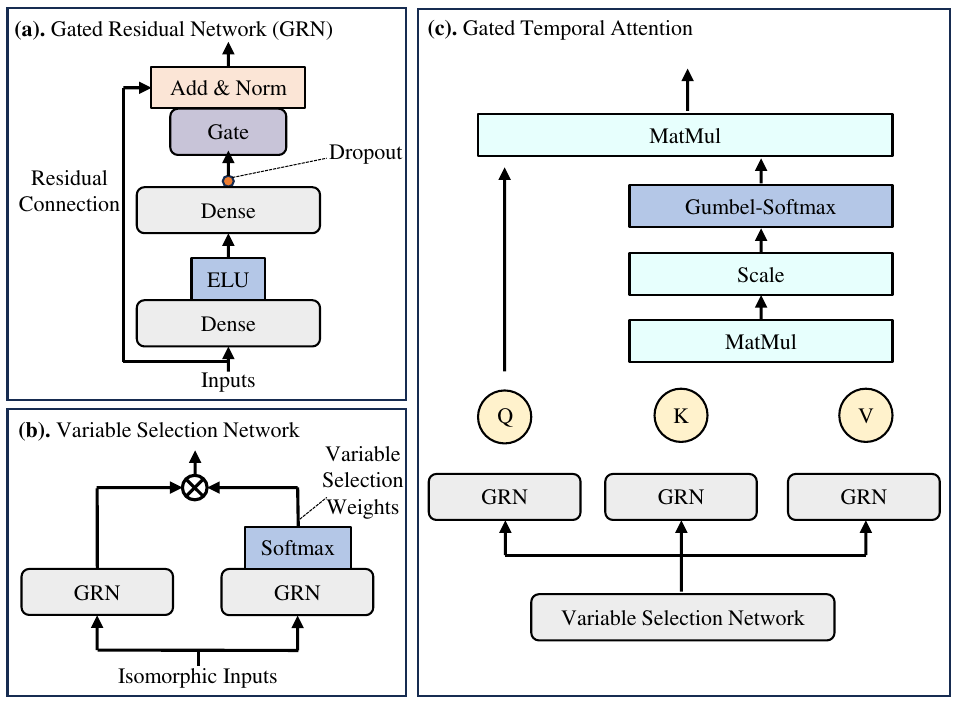}
	\caption{Schematic of some module structures. (a). Gated residual network, (b). Variable selection network and (c). Gated temporal attention.}
	\label{module}
\end{figure}

\autoref{module} illustrates the structure of three building blocks of the proposed model. First, as shown in \autoref{module}(a), we introduce Gated Residual Network (GRN) as a building block to allow the model to flexibly handle nonlinear relationships when needed. Unlike traditional networks that use simple linear layers, we choose GRNs in order to provide the network with more powerful understanding capabilities. GRNs have more sophisticated gating mechanisms and nonlinear activation functions compared to linear layers, and can adaptively choose nonlinear processing in response to fluctuations in charging demand. The specific calculation process of GRN is shown in \autoref{GRN}.

\begin{equation}
	\label{GRN}
	\begin{aligned}
		\bm{\eta}_1 ={\rm ELU}(\bm{W}_{\eta_1}\bm{x}+\bm{b}_{\eta_1})\\
		\bm{\eta}_2 =\bm{W}_{\eta_2}\bm{\eta}_1+\bm{b}_{\eta_2}\\
		{\rm GRN}(\bm{x}) = {\rm{Add \& Norm}}(\bm{x}, {\rm GLU}(\bm{\eta}_2))
	\end{aligned}
\end{equation}
where $\bm{x}$ denotes input matrix, $\bm{W}_{\eta_1}, \bm{W}_{\eta_2} , \bm{b}_{\eta_1}, \bm{b}_{\eta_2}$ are coefficient matrices, $\bm{\eta_1}, \bm{\eta_2}$ are intermediate layers, and ELU is the Exponential Linear Unit activation function \citep{ELUarticle}. Furthermore, Gated Linear Unit (GLU) is used as gate mechanism to flexibly handle both necessary and non-necessary information for given data, shown as \autoref{GLU}.

\begin{equation}
	\label{GLU}
	{\rm GLU}(\bm{\eta} _2) = (\bm{W}_1^{{\rm GLU}} \bm{\eta} _2+\bm{b}_1^{{\rm GLU}})\odot \sigma (\bm{W}_2^{{\rm GLU}}\bm{\eta} _2+\bm{b}_2^{{\rm GLU}})
\end{equation}
where $\bm{W}_1^{{\rm GLU}}, \bm{W}_2^{{\rm GLU}}, \bm{b}_1^{{\rm GLU}}, \bm{b}_2^{{\rm GLU}}$ are coefficient matrices, $\odot$ denotes the element-wise Hadamard product, and $\sigma(.)$ stands for the sigmoid activation function. During training, dropout is applied after the second dense layer, i.e., to $\eta_2$ in \autoref{GRN}.

\begin{figure}[htbp]
	\centering
	\includegraphics[width=4.5in]{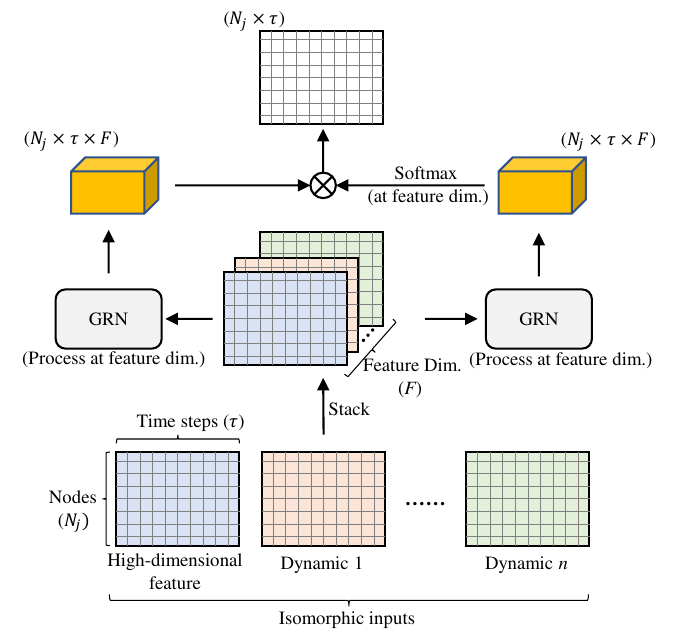}
	\caption{Detailed operation of the proposed variable selection network.}
	\label{SeleNet}
\end{figure}

As shown in \autoref{module}(b) and \autoref{SeleNet}, we propose a variable selection network to incorporate dynamic information. Auxiliary information and historical demand data used for urban EV charging demand prediction are inputted. The specific contribution of variables in each city area is determined by the network, i.e., variable selection network enables to emphasize variables that are critical to the prediction problem and ignore variables that are not independently at each node. The stacked input data $\bm{\xi}\in \mathbb{R} ^{N_j \times \tau \times F} $ (including high-dimensional features $\bm{H}$, and other dynamics) are processed independently by two GRNs at feature dimension. A softmax function after one of the GRNs determines the variable selection weights, which are then multiplied with the embedded features. The calculation process of the variable selection network is shown in \autoref{VSN}. Note that we do not operate on the node dimension and the timestep dimension, which means that attention can be allocated independently in spatio-temporal terms. In other words, the importance of dynamic features can be differentially determined.

\begin{equation}
	\label{VSN}
	\begin{aligned}
		\bm{\xi}_1 ={\rm GRN}(\bm{\xi})\\
		\bm{v}_1 ={\rm Softmax}({\rm GRN}(\bm{\xi}))\\
		\bm{\xi}_2 = \bm{v}_1\bm{\xi}_1
	\end{aligned}
\end{equation}
where $\bm{\xi}_1 \in \mathbb{R} ^{N_j \times \tau \times F}$ is the embedded features after the first step calculation, $\bm{v}_1 \in \mathbb{R} ^{N_j \times \tau \times F}$ is the variable selection weights, and $\bm{\xi}_2 \in \mathbb{R} ^{N_j \times \tau}$ denotes the output of variable selection network, which combines multiple features. $N_j$ denotes the number of city areas, $\tau$ denotes the number of time steps, and $F$ stands for the number of features.

Transformer utilizes self-attention mechanism to model sequences, breaking through the limitation that recurrent neural networks cannot be computed in parallel, and becoming a research hotspot for time series tasks. Considering the historical correlation and slight fluctuations in charging demand data, we adopt the framework of Transformer encoder and improve it by utilizing GRN and Gated Attention Network. Similarly to self-attention, we process the output of variable selection network (i.e., $\bm{\xi}_2$) at time step dimension independently with three GRNs to obtain the query, key, and value matrix $\bm Q$, $\bm K$, and $\bm V$.

The attention mechanism assigns weights to each input time step, and even irrelevant units are assigned small weights. This leads to two problems: 1) relevant units are assigned insufficient attention weights, and 2) noise units without contributions are assigned weights, leading to performance degradation. Considering that in long charging demand sequences not every interval is sufficiently helpful for prediction, we develop gated temporal attention, in which we perform a Scaled Dot-Product Attention operation, but Gumbel-Softmax is used as weight activation function instead of Softmax, as shown in \autoref{self atten} and \autoref{module}(c).

\begin{equation}
	\label{self atten}
	{A_{T}}(\bm{Q},\bm{K},\bm{V})={\rm Gumbel}\mbox{-}{\rm Softmax}(\frac{\bm{Q}\bm{K}^{T}}{\sqrt{d_{k}}} )\bm{V}
\end{equation}
where $A_{T}$ is the output of the improved attention operation, and $\sqrt{d_{k}}$ acts as a scaling factor.

Gumbel-Softmax acts as binary gates and achieve sparse attention on long sequences. It provides the model with the ability to perform gradient descent which binary gates cannot. Gumbel-Softmax can be formulated as \autoref{Gumbel-Softmax}.

\begin{equation}
	\label{Gumbel-Softmax}
	\hat{\alpha}  _i = \frac{{\rm exp}(({\rm log}(\alpha_i)+\epsilon _i)/\varpi)  }{ {\textstyle \sum_{j}^{}{\rm exp}(({\rm log}(\alpha_j)+\epsilon _j)/\varpi)} }
\end{equation}
where $\hat{\alpha}  _i$ is the attention weight of $i$-th unit, $\alpha_i$ is $i$-th position of the input vector, $\epsilon _i$ is a random sample from Gumbel$(0, 1)$, and $\varpi$ denotes the temperature coefficient, in other words a scaling factor.

Similar to Transformer encoder, the proposed data processing process is formulated as \autoref{encoder}.

\begin{equation}
	\label{encoder}
	\begin{aligned}
		\bm{\psi }_1 = {\rm{GTA}}(\bm{\psi })\\
		\bm{\psi }_2 = {\rm{Add \& Norm}}(\bm{\psi }_1, \bm{\psi })\\
		\bm{\psi }_3 = {\rm{GRN}}(\bm{\psi }_2)\\
		\bm{\psi }_4 = {\rm{Add \& Norm}}(\bm{\psi }_2, \bm{\psi }_3)\\
	\end{aligned}
\end{equation}
where $\bm{\psi }$ is input matrix, which is equal to $\bm{\xi}_2$ in \autoref{VSN}, $\bm{\psi} _i$ denotes the output of the $i$-th step, and $\rm GTA$ stands for gated temporal attention as \autoref{self atten}. Note that $\bm{\psi }_4$ can be consider as an input to the next encoder layer, and the operation of encoder is repeated $N_E$ times.

Finally, a dense layer is used to decode the information and output the predictions.

\section{Experiments}
\label{experiments}
In this section, we first present the data description and the specific setup of the experiment. Then, we compare the proposed model with current state-of-the-art models to empirically illustrate the performance of CityEVCP. The model generalizability is also illustrated. Furthermore,we conduct ablation experiments to show that each module and each factor can provide a contribution to improving the prediction accuracy. Finally, the impact of hyperparameters is analyzed experimentally.

\subsection{Data Description}
The proposed model CityEVCP is validated on a benchmark dataset collected from June 19 to July 18, 2022 in Shenzhen, China. We divide the city into areas and update the occupancy rate of the counted charging piles within the area every 5 minutes. Consequently we can obtain 8640 timestamps for the month. The areas division and the number of charging piles included in each area are shown in this link\footnote{https://github.com/IntelligentSystemsLab/ST-EVCDP}. We divide the training set, validation set and test set by 6:1:3 in chronological order, i.e., 19 June to 6 July for training, 7 July to 9 July for validation and 10 July to 18 July for test. A total of 450,175 POI data are collected from Amap\footnote{https://lbs.amap.com/}. POIs include a total of 14 categories such as restaurant, shopping, scenic areas, etc. Temperature data is also collected from a weather station in Shenzhen, which has a sampling interval of 30 minutes. We use linear interpolation to enable a 5-minute interval in the temperature data, making it the same as the demand data.

\subsection{Experimental Setup}
\label{setup}
The hyperparameters are chosen after a detailed search experiment and we provide hyperparameter analysis in \autoref{Hyperparameter}. The final experimental setup we chose is as follows. First, the lookback window size $\tau$ is set to 12, i.e. we provide the model with 60 minutes of historical information. Second, in K-means clustering, we cluster the area into 10 classes. Third, the temperature coefficient $\varpi$ is set to 1.5 to not only provide sparse attention to temporal information, but also to avoid focusing attention on one-hot vector. Fourth, the number of encoder components $N_E$ is set to 2. Fifth, the maximum epoch numbers of training process is set to 2000, and model training will early stop if the loss of the validation set does not decrease within 50 epochs. Sixth, Mean Square Error (MSE) is used as the loss function for all comparison models. Seventh, we use Adam as an optimizer with a mini-batch size of 512. Seventh, we employ four metrics, i.e., Root Mean Squared Error (RMSE), R-Square Coefficient ($\rm{R}^{2}$), Relative Absolute Error (RAE), and Mean Absolute Error (MAE) to illustrate the performance of our proposed model. Finally, all experiments in this study are accomplished on a Windows personal computer containing a NVIDIA RTX 4000 GPU. We share the code for our proposed model CityEVCP in this link.\footnote{https://github.com/kuanghx3/CityEVCP}.

\subsection{Comparison Experiment}
In this subsection, the proposed model CityEVCP will be compared with representative models to illustrate the state-of-the-art performance. The models for comparison includes 3 sequential neural networks, 2 spatial neural networks and 5 spatio-temporal neural networks, which are described in detail as follows:

\begin{itemize}
	
	\item Long Short-Term Memory (LSTM) \citep{hochreiter1997long, wang2023short}: An improved recurrent neural network that can efficiently process temporal information.
	
	\item State Space Model (SSM) \citep{guefficiently}: A hotspot model for efficient sequence modeling by capturing key information in sequences through hidden states.
	
	\item Temporal Fusion Transformer (TFT) \citep{lim2021temporal}: A model combining high-performance multiview prediction with interpretable temporal dynamic insights based on a Transformer architecture.
	
	\item Graph Convolutional Network (GCN) \citep{kipf2016semi}: A convolutional neural network that can act directly on graphs and utilize their structural information.
	
	\item General Hypergraph Neural Network (HGNN) \citep{gao2022hgnn+}: A hypergraph convolution scheme to learn generalized data representations for a variety of tasks.
	
	\item TGCN \citep{zhao2019t}: A neural network-based traffic prediction method combining graph convolutional networks and gated recurrent units.
	
	\item AGGRU \citep{zhang2023tmfo}: A graph convolutional gated recurrent network. Unlike existing deep learning models, linear operations in gated recurrent neural networks are replaced by graph convolution operations.
	
	\item BEGAN \citep{xu2023air}: An air traffic demand prediction network combining long short-term memory and graph attention mechanisms.
	
	\item FCSTGNN \citep{wang2024fully}: A fully-connected spatial-temporal graph neural network that captures integrated spatio-temporal dependencies in multivariate time series data by graph convolution operations.
	
	\item HSTGCN \citep{wang2023predicting}: A heterogeneous spatio-temporal graph convolutional network combining graph convolution and gated recurrent units for predicting EV charging demand at different spatio-temporal resolutions. In particular, it considers POI classification through separate fully connected networks.
	
\end{itemize}

\begin{table}[htbp]
	\centering
	\caption{Performance of the compared models.}
	\begin{threeparttable}
	\begin{tabular}{c|ccccc|ccccc}
		\toprule
		Metrics ($\times 10^{-2}$) & \multicolumn{5}{c|}{RMSE}             & \multicolumn{5}{c}{$ \rm {R}^2$} \\
		Model & 15min & 30min & 45min & 60min & Average & 15min & 30min & 45min & 60min & Average \\
		\midrule
		LSTM  & 3.49  & 4.76  & 5.73  & 6.58  & 5.14  & 89.85 & 82.11 & 74.79 & 67.17 & 78.48 \\
		SSM   & 3.51  & 4.80   & 5.82  & 6.70   & 5.21  & 89.94 & 82.2  & 74.70  & 66.97 & 78.45 \\
		TFT   & 3.47  & 4.60  & 5.48  & 6.18  & 4.93  & 88.14 & 81.40 & 73.68 & 66.60 & 77.46 \\
		GCN   & 3.73  & 4.95  & 5.88  & 6.70  & 5.31  & 86.35 & 77.82 & 70.15 & 62.10 & 74.11 \\
		HGNN  & 3.24  & 4.60  & 5.58  & 6.45  & 4.97  & 90.11 & 81.62 & 74.27 & 66.29 & 78.07 \\
		TGCN  & 3.59  & 4.47  & 5.73  & 6.53  & 5.08  & 87.14 & 83.98 & 71.60 & 63.72 & 76.61 \\
		AGGRU & 3.04  & 4.80  & 5.50  & 6.39  & 4.93  & 92.13 & 79.07 & 76.71 & 69.04 & 79.24 \\
		BEGAN & 2.95  & 4.41  & 5.45  & 6.40  & 4.80  & 92.35 & 83.85 & 75.65 & 65.00 & 79.21 \\
		FCSTGNN & 3.07  & 4.72  & 5.39  & 6.26  & 4.86  & 90.05 & 79.83 & 74.10 & 67.09 & 77.77 \\
		HSTGCN & 2.88  & 4.35  & 5.38  & 6.27  & 4.72  & 92.39 & 83.92 & 76.60 & 68.86 & 80.44 \\
		CityEVCP & \textbf{2.74} & \textbf{4.18} & \textbf{5.16} & \textbf{5.96} & \textbf{4.51} & \textbf{92.91} & \textbf{84.35} & \textbf{77.14} & \textbf{69.63} & \textbf{81.01} \\
		\midrule
		Metrics ($\times 10^{-2}$) & \multicolumn{5}{c|}{RAE}              & \multicolumn{5}{c}{MAE} \\
		Model & 15min & 30min & 45min & 60min & Average & 15min & 30min & 45min & 60min & Average \\
		\midrule
		LSTM  & 13.20 & 18.35 & 22.62 & 26.76 & 20.23 & 1.83  & 2.55  & 3.14  & 3.71  & 2.81 \\
		SSM   & 12.37 & 17.74 & 22.20  & 26.32 & 19.66 & 1.72  & 2.46  & 3.08  & 3.65  & 2.73 \\
		TFT   & 14.17 & 18.77 & 23.01 & 26.39 & 20.59 & 1.97  & 2.60  & 3.19  & 3.66  & 2.86 \\
		GCN   & 14.97 & 20.14 & 24.37 & 28.33 & 21.95 & 2.08  & 2.79  & 3.38  & 3.93  & 3.04 \\
		HGNN  & 12.05 & 17.85 & 22.34 & 26.59 & 19.71 & 1.67  & 2.48  & 3.10  & 3.69  & 2.73 \\
		TGCN  & 14.39 & 16.34 & 23.62 & 27.52 & 20.47 & 2.00  & 2.27  & 3.28  & 3.82  & 2.84 \\
		AGGRU & 10.35 & 19.47 & 21.08 & 25.43 & 19.08 & 1.44  & 2.70  & 2.92  & 3.53  & 2.65 \\
		BEGAN & 10.15 & 16.64 & 21.79 & 26.95 & 18.88 & 1.41  & 2.31  & 3.02  & 3.74  & 2.62 \\
		FCSTGNN & 13.25 & 19.90 & 23.03 & 26.80 & 20.75 & 1.84  & 2.76  & 3.19  & 3.72  & 2.88 \\
		HSTGCN & 10.15 & 16.42 & 20.99 & 25.23 & 18.20 & 1.41  & 2.28  & 2.91  & 3.50  & 2.52 \\
		CityEVCP & \textbf{9.86} & \textbf{16.13} & \textbf{20.52} & \textbf{24.41} & \textbf{17.73} & \textbf{1.37} & \textbf{2.24} & \textbf{2.85} & \textbf{3.39} & \textbf{2.46} \\
		\bottomrule
	\end{tabular}%
	\begin{tablenotes}
		\footnotesize
		\item \textbf{Bold} denotes the best results. Same below.
	\end{tablenotes}
	\end{threeparttable}
	\label{tab:all}%
\end{table}%

In \autoref{tab:all}, we demonstrate the performance of the above comparison models and CityEVCP in four prediction intervals (i.e., 15, 30, 45, and 60 minutes). The proposed model is the best in each prediction interval and on each metric. Compared to ten representative methods, our proposed method CityEVCP improves about 9.59\% in RMSE, 10.90\% in RAE, and 10.91\% in MAE on average, and achieves the best fit in $\rm R^2$. Moreover, CityEVCP outperforms the ten comparison models by 18.50\%, 10.41\%, 8.31\% and 7.97\% on average in 15, 30, 45, and 60 minutes prediction tasks, respectively. CityEVCP performs well on relatively long-term prediction, which is a challenge in prediction problems, demonstrating the contribution of temporal modules (e.g., improved transformer encoder) and multi-source heterogeneous information to improve the accuracy of EV charging demand prediction. Specifically, we consider the same two time-series effects, price and temperature, in previously proposed multivariate time-series convolutional model, FCSTGNN, and in comparison, CityEVCP improves by 12.08\% over FCSTGNN on average. Moreover, compared to the previous EV charging demand prediction method, HSTGCN, which considers points of interest in separate fully connected networks, our proposed model improves by 3.19\%. This demonstrates that the effective introduction of external information such as the areas' attributes and weather, as well as the improvement of the network structure can enhance the prediction accuracy of citywide EV charging demand.

\begin{figure}[htbp]
	\centering
	\includegraphics[width=6.5in]{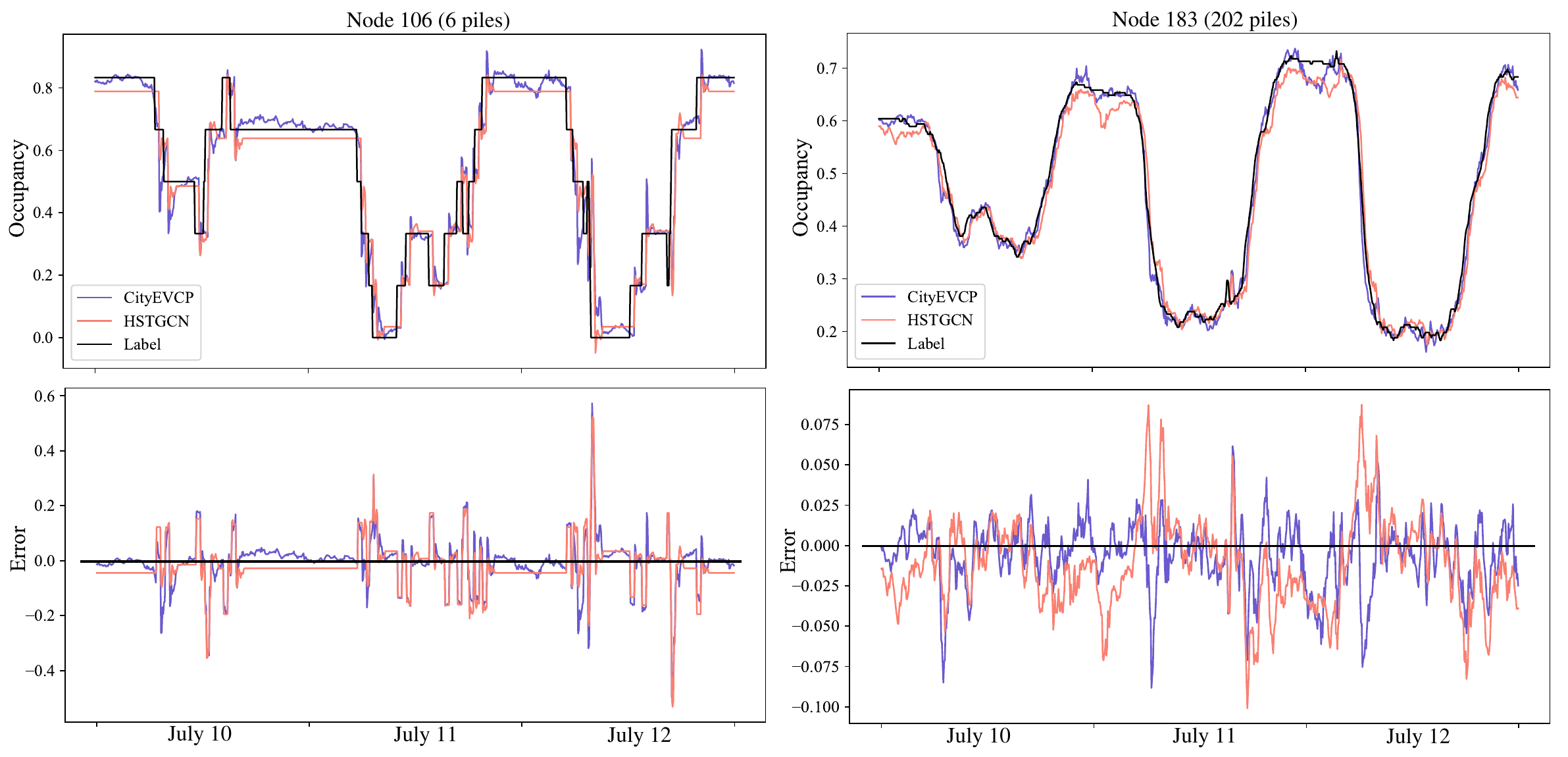}
	\caption{Lineplots of 60-min prediction results and errors for two example areas.}
	\label{plot}
\end{figure}

To further illustrate the performance of the proposed model, we show the prediction results for two typical areas in \autoref{plot}. The prediction results of CityEVCP and HSTGCN and true area charging demand (i.e., occupancy) are shown. Although the proposed model CityEVCP does not perform perfectly at all times, it outperforms HSTGCN overall. The errors between two methods and the true label are also shown, in order to intuitively illustrate the performance of the model. Node 106 is a suburb located in Bao'an District, Shenzhen, with only six charging piles. Due to too few charging piles in the area, the demand for charging shows stability at some times of a day. CityEVCP is more willing to adjust than HSTGCN, even though the true labels do not change, which results in CityEVCP's prediction accuracy being closer to the true values. This may be due to the ability of the proposed model to take into account dynamic auxiliary information, and reduce the inertia and lags. Node 183 is located in a residential area in Longgang District, Shenzhen and has 202 charging piles. CityEVCP has a smaller error than HSTGCN, illustrating its good performance in regular areas. CityEVCP copes well with small fluctuations in daily data because the model can enable sparse temporal attention and fully account for urban areas and dynamic influences. In contrast, HSTGCN tends to amplify data fluctuations and develop large prediction errors.

In comparison models, firstly, GCN that only considers spatial adjacencies and lacks consideration of temporal factors is the worst. Second, complex models perform better than simple ones, because complex models can model nonlinear spatio-temporal relationships accurately. This is why current research favors delicate and complex networks, even though they may require more storage space and calculation. Thirdly, the consideration of influences on EV charging demand somehow brings about improvements in model performance. HGNN and HSTGCN, which incorporate area attribute information, have relatively outstanding performance in comparison models, illustrating that area attributes can help model spatial features from high dimensions and improve prediction accuracy by leveraging similar nodes. TFT and FCSTGNN, which combine dynamic influences, outperform models of equal volume in long-time prediction tasks (e.g.,45min and 60min), illustrating the important role of dynamic auxiliary information in enhancing the model's forward-looking capability.

We conduct additional experiments to show that our model does not just ﬁt on a speciﬁc dataset. We randomly drop 20\% of the 247 areas to create different geographic contexts in the experiments. To avoid randomness, the experiments are repeated five times and their results are averaged. On these new datasets, we selected one temporal model (i.e., LSTM) and three spatio-temporal models (i.e., BEGAN, FCSTGNN, and HSTGCN) as comparative models to demonstrate the performance of our proposed model. The results and average results for the 4 prediction intervals are showed in \autoref{tab:drop}. In the additional experiments, the proposed model CityEVCP improves by 11.94\% in RMSE, 12.95\% in RAE and 13.50\% in MAE on average compared to the comparison model, and achieves the best fit in $\rm R^2$.

\begin{table}[htbp]
	\centering
	\caption{Experiment results of random node dropping.}
	\begin{tabular}{c|ccccc|ccccc}
		\toprule
		Metrics ($\times 10^{-2}$) & \multicolumn{5}{c|}{RMSE}             & \multicolumn{5}{c}{$ \rm {R}^2$} \\
		Model & 15min & 30min & 45min & 60min & Average & 15min & 30min & 45min & 60min & Average \\
		\midrule
		LSTM  & 4.38  & 5.42  & 6.27  & 7.02  & 5.77  & 83.30 & 75.86 & 68.48 & 60.96 & 72.15 \\
		BEGAN & 2.89  & 4.45  & 5.46  & 6.34  & 4.78  & 92.60 & 82.54 & 75.84 & 68.60 & 79.89 \\
		FCSTGNN & 3.02  & 4.37  & 5.36  & 6.16  & 4.73  & 90.84 & 82.35 & 74.02 & 66.47 & 78.42 \\
		HSTGCN & 3.16  & 4.56  & 5.54  & 6.24  & 4.87  & 90.62 & 82.01 & 74.56 & 68.98 & 79.04 \\
		CityEVCP & \textbf{2.69} & \textbf{4.08} & \textbf{5.05} & \textbf{5.83} & \textbf{4.41} & \textbf{92.81} & \textbf{84.50} & \textbf{77.09} & \textbf{69.63} & \textbf{81.01} \\
		\midrule
		Metrics ($\times 10^{-2}$) & \multicolumn{5}{c|}{RAE}              & \multicolumn{5}{c}{MAE} \\
		Model & 15min & 30min & 45min & 60min & Average & 15min & 30min & 45min & 60min & Average \\
		\midrule
		LSTM  & 18.44 & 22.80 & 26.64 & 30.27 & 24.54 & 2.52  & 3.12  & 3.64  & 4.14  & 3.36 \\
		BEGAN & 10.22 & 17.64 & 22.11 & 26.30 & 19.07 & 1.40  & 2.41  & 3.02  & 3.59  & 2.61 \\
		FCSTGNN & 12.84 & 18.69 & 23.52 & 27.50 & 20.64 & 1.75  & 2.54  & 3.20  & 3.74  & 2.81 \\
		HSTGCN & 11.81 & 17.99 & 22.48 & 25.77 & 19.51 & 1.62  & 2.46  & 3.07  & 3.52  & 2.67 \\
		CityEVCP & \textbf{10.16} & \textbf{16.27} & \textbf{21.03} & \textbf{24.75} & \textbf{18.05} & \textbf{1.38} & \textbf{2.21} & \textbf{2.85} & \textbf{3.36} & \textbf{2.45} \\
		\bottomrule
	\end{tabular}%
	\label{tab:drop}%
\end{table}%

\subsection{Ablation experiment}
A comprehensive ablation experiment is designed to validate the contribution of each part of our proposed model CityEVCP. We remove module (a), (b), and (c), price information, temperature information, and variable selection networks, respectively. Specifically, we also replace Gumbel-Softmax with Softmax in the gate temporal attention module to demonstrate its usefulness. In \autoref{tab:ablation}, we present the results of ablation experiments, which are conducted in 4 different intervals (i.e., 15, 30, 45, and 60 min). The full model, CityEVCP, improves about 10.04\% in RMSE, 15.85\% in RAE, and 15.85\% in MAE on average. In the 60-minute prediction, the proposed model CityEVCP is weakly worse than “\# Gumbel-Softmax” in $\rm R^2$, but better in the other three metrics. We still argue that the ablation experiment can be passed because its overall performance is not bad. Different modules and information contribute differently to improving prediction accuracy. First, Module (b) (i.e. adjacency graph fusion module) contributes the most to the prediction, demonstrating the importance of learning adjacency features. Second, Module (c) (i.e. temporal feature fusion module) is helpful in improving long-time prediction accuracy, and removing Module (c) sharply deteriorates the results as the prediction interval increases. Besides, time-series information, such as prices and temperatures, both contribute to the accuracy of long-term prediction. Third, module (a) enables the model to effectively capture the dynamic changes of the areas with same land use, thus improving prediction accuracy. Fourth, the variable selection network helps to determine the weights among the temporal variables, and the gated temporal attention allows a strong focus on important temporal features, both of which provide a non-negligible boost.

\begin{table*}[htbp]
	\centering
	\caption{Results of ablation experiment in four prediction intervals.}
	\begin{threeparttable}
		\begin{tabular}{c|ccccc|ccccc}
			\toprule
			Metrics ($\times 10^{-2}$) & \multicolumn{5}{c|}{RMSE}             & \multicolumn{5}{c}{$\rm R^2$} \\
			Model & 15min & 30min & 45min & 60min & Average & 15min & 30min & 45min & 60min & Average \\
			\midrule
			\# Module (c) & 3.23  & 4.95  & 5.90  & 6.73  & 5.20  & 90.03 & 76.48 & 67.51 & 57.23 & 72.81 \\
			\# Module (b) & 4.81  & 4.95  & 5.68  & 6.41  & 5.46  & 71.11 & 75.55 & 71.43 & 64.09 & 70.54 \\
			\# Module (a) & 3.39  & 4.52  & 5.58  & 6.20  & 4.92  & 88.90 & 81.86 & 70.83 & 66.20  & 76.95 \\
			\# price & 3.65  & 4.78  & 5.69  & 6.46  & 5.15  & 85.83 & 78.49 & 71.34 & 63.48 & 74.78 \\
			\# temperature & 3.37  & 4.60  & 5.56  & 6.41  & 4.98  & 88.37 & 80.94 & 72.69 & 64.65 & 76.66 \\
			\# Var. Sele. & 3.17  & 4.70  & 5.53  & 6.39  & 4.95  & 89.44 & 79.42 & 73.36 & 65.15 & 76.84 \\
			\# Gumbel-Softmax & 2.80  & 4.19  & 5.19  & 5.98  & 4.54  & 92.40 & 84.28 & 76.59 & \textbf{69.67} & 80.73 \\
			Full  & \textbf{2.74} & \textbf{4.18} & \textbf{5.16} & \textbf{5.96} & \textbf{4.51} & \textbf{92.91} & \textbf{84.35} & \textbf{77.14} & 69.63 & \textbf{81.01} \\
			\midrule
			Metrics ($\times 10^{-2}$) & \multicolumn{5}{c|}{MAE}              & \multicolumn{5}{c}{RAE} \\
			Model & 15min & 30min & 45min & 60min & Average & 15min & 30min & 45min & 60min & Average \\
			\midrule
			\# Module (c) & 12.49 & 20.49 & 25.07 & 29.46 & 21.88 & 1.73  & 2.84  & 3.48  & 4.09  & 3.03 \\
			\# Module (b) & 22.60 & 20.80 & 23.60 & 27.30 & 23.58 & 3.14  & 2.89  & 3.27  & 3.79  & 3.27 \\
			\# Module (a) & 14.16 & 18.47 & 23.67 & 26.38 & 20.67 & 1.96  & 2.56  & 3.28  & 3.66  & 2.87 \\
			\# price & 16.27 & 20.57 & 24.21 & 27.82 & 22.22 & 2.26  & 2.85  & 3.36  & 3.86  & 3.08 \\
			\# temperature & 14.20 & 19.05 & 23.16 & 27.33 & 20.93 & 1.97  & 2.64  & 3.21  & 3.79  & 2.90 \\
			\# Var. Sele. & 13.41 & 20.14 & 23.13 & 27.13 & 20.95 & 1.86  & 2.79  & 3.21  & 3.76  & 2.91 \\
			\# Gumbel-Softmax & 10.30 & 16.38 & 21.05 & 24.73 & 18.11 & 1.43  & 2.27  & 2.92  & 3.43  & 2.51 \\
			Full  & \textbf{9.86} & \textbf{16.13} & \textbf{20.52} & \textbf{24.41} & \textbf{17.73} & \textbf{1.37} & \textbf{2.24} & \textbf{2.85} & \textbf{3.39} & \textbf{2.46} \\
			\bottomrule
		\end{tabular}%
		\begin{tablenotes}
			\footnotesize
			\item Note that "\#" denotes "without", and "Var. Sele." denotes Variable selection network.
			\item "\# Gumbel-Softmax" means replace Gumbel-Softmax with Softmax in Gated temporal attention module.
			
		\end{tablenotes}
	\end{threeparttable}
	\label{tab:ablation}%
\end{table*}%

\begin{figure*}[tbp]
	\centering
	\includegraphics[width=6.5in]{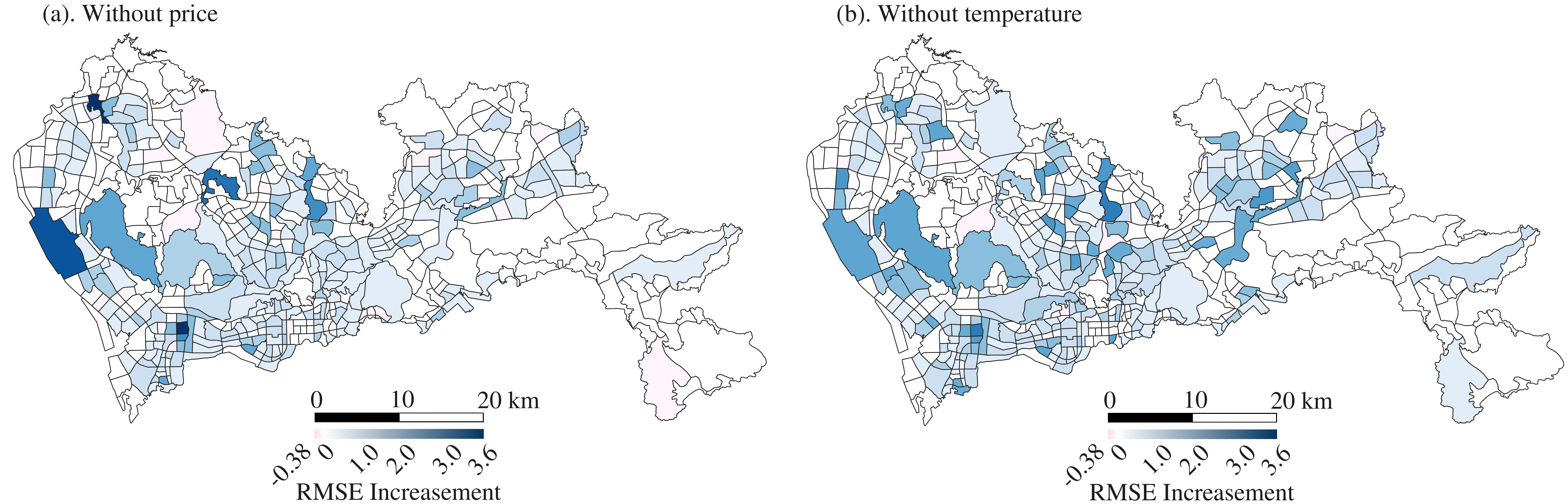}
	\caption{Heatmap of 60-min prediction performance declines by removing factors. Errors are measured in RMSE.}
	\label{dynamic_ablation}
\end{figure*}

Furthermore, in \autoref{dynamic_ablation}, we visualize the performance decline in each area in long-term prediction (i.e., 60 minutes) due to the removal of dynamic factors. The removal of price and temperature has led to a decline in overall urban EV charging demand prediction performance, which confirms the correlation between dynamic factors and charging demand. In our case, price information contributes more to prediction accuracy than temperature information, which can be concluded in \autoref{tab:ablation} and \autoref{dynamic_ablation}. In addition, different dynamic information contributes differently to the prediction accuracy in the city. Price information contributes strongly to the residential areas (e.g., western and northwestern Shenzhen), but weakly contributes, or even brings a weak negative effect to suburban areas (e.g., eastern and northern tourist areas). In contrast, temperature information makes a positive contribution to the suburban and tourist areas, as well as to other areas of the city. This is in line with common sense. Residents' charging in their residential areas is sensitive to price, while leisure and recreational activities are affected by the weather easily. It shows that our model can effectively learn the pattern of urban residents' travel behavior influenced by dynamic factors.

\subsection{Spatial correlation analysis}

\begin{figure}[htbp]
	\centering
	\includegraphics[width=6.5in]{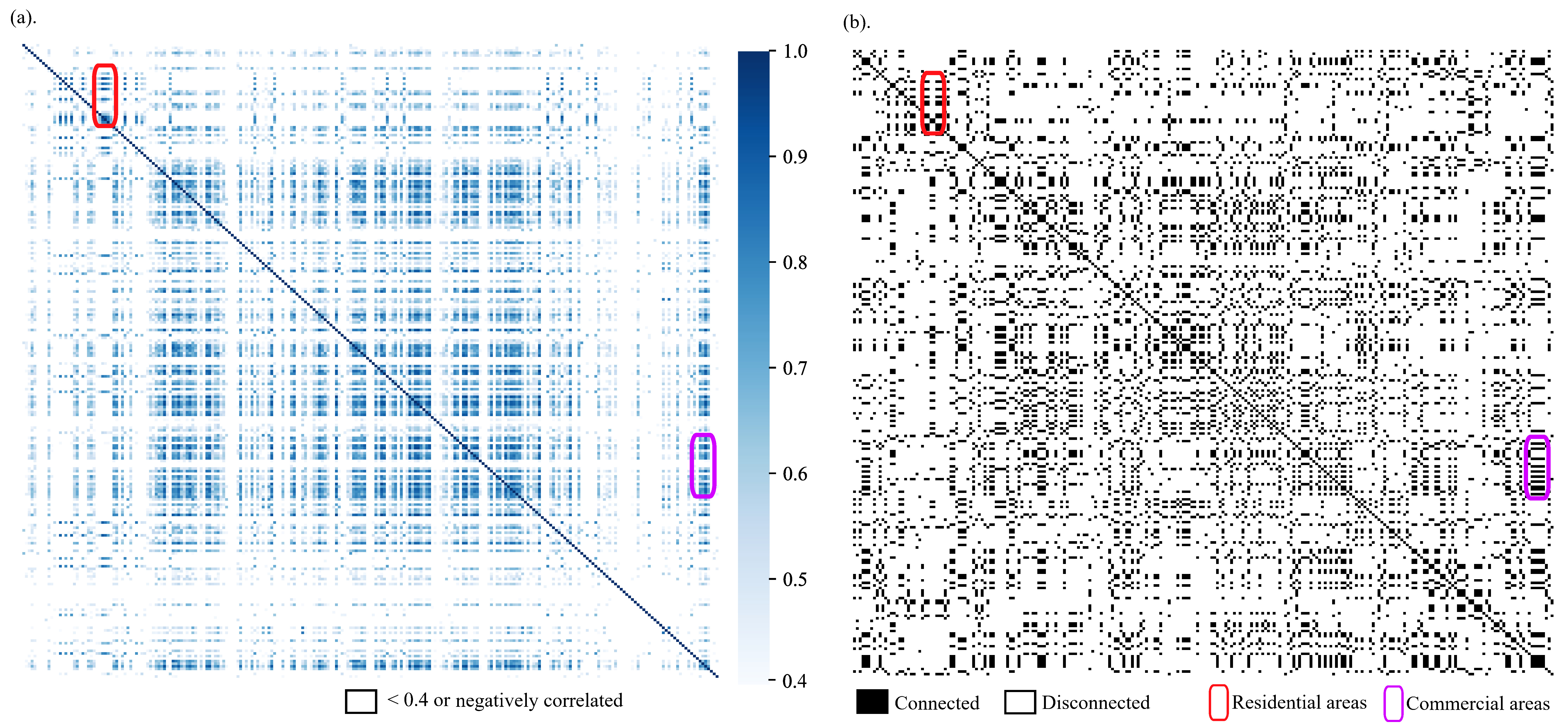}
	\caption{(a).Heat map of Pearson's correlation coefficient; (b). Connections in our hypergraph.}
	\label{corr}
\end{figure}

To illustrate the effectiveness of our geo-based clustering method, we show the true correlation of 247 areas' occupancies and the connectivity relationship obtained by our clustering method in \autoref{corr}. \autoref{corr}(a) illustrates the areas that show a strong positive correlation. We consider a relatively strong correlation when the Pearson correlation coefficient between the occupancy data of two areas is greater than or equal to 0.4. \autoref{corr}(b) illustrates the relations connected by hyperedges in our hypergraph. A black square indicates that the two areas are connected by a hyperedge, and conversely white one indicates that they are not connected. The two images are similar in many places, which shows that our clustering method can effectively capture spatial data correlation. To give concrete examples, the red box shows some residential areas in Luohu District, Shenzhen, and the purple box shows some commercial areas in Futian District, Shenzhen. Areas with the same land use type have similar patterns of changing charging demand as shown in \autoref{corr}(a). Corresponding to that, our approach makes good use of POIs to cluster areas with similar land use attributes and construct hyperedge connectivity relationships.

\subsection{Hyperparameter analysis}
\label{Hyperparameter}

\begin{figure*}[htbp]
	\centering
	\includegraphics[width=6in]{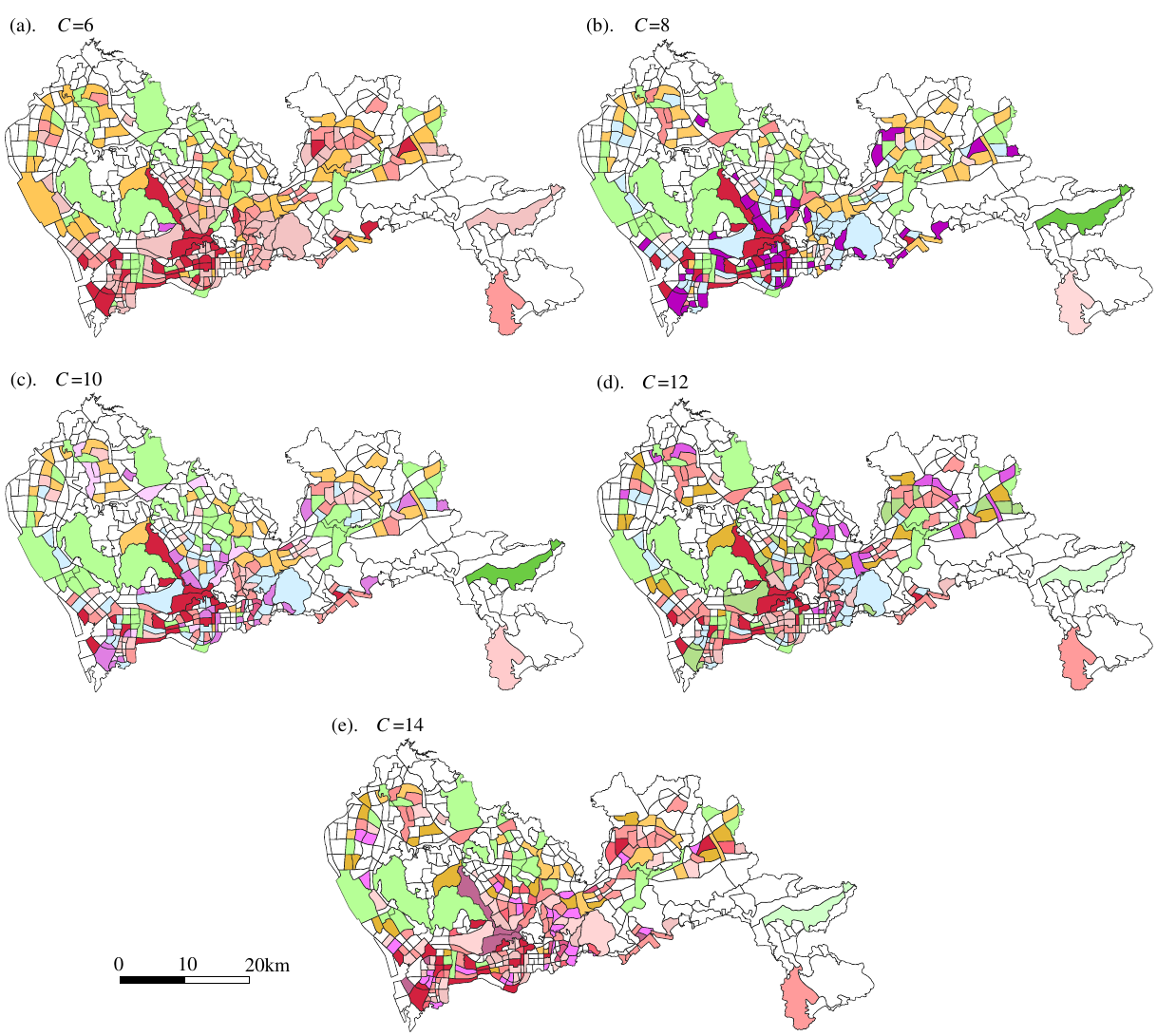}
	\caption{Impact of the clustering number on the results.}
	\label{clustering}
\end{figure*}

\begin{figure}[htbp]
	\centering
	\includegraphics[width=6.5in]{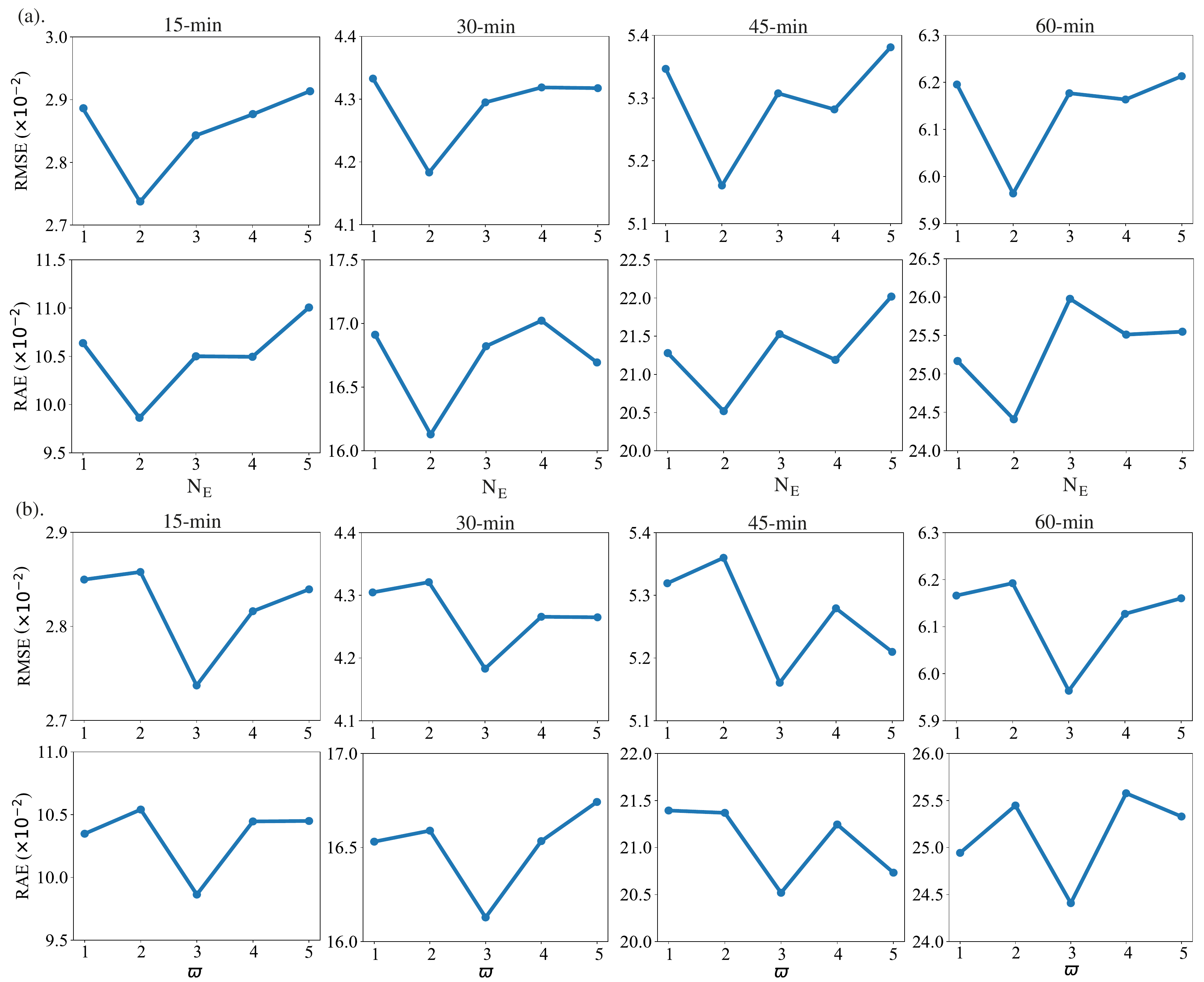}
	\caption{Impact of hyperparameters (i.e., $N_E$ and $\varpi$) on prediction accuracy.}
	\label{params}
\end{figure}

In this subsection, we discuss the impact of hyperparameters (i.e., number of adaptive areas clusters $C$, number of encoder components $N_E$ and temperature coefficient of Gumbel-Softmax $\varpi$). First, as shown in \autoref{clustering}, despite $C$ being set differently, the area adaptive clustering roughly matches reality. The southern urban built-up area, developing north-western, north-eastern and central parts, and forests, reservoirs and city parks are clustered separately. However, when $C$ is too small (e.g., $C = 6$), the method has difficulty in distinguishing between urban sub-centers and suburban tourist areas.  Besides, when $C$ is too large, the method tends to cut the city into too many patches, e.g., as many as six types of functional areas in the southern CBD area when $C=14$, which is unfavorable for subsequent hypergraph studies. Second, we determine the appropriate hyperparameters by searching experiment, as shown in \autoref{params}. Our experiment results demonstrate the robustness of the proposed model CityEVCP, as it is insensitive to the change of hyperparameters. $N_E$ affects the depth of the network. Networks that are too shallow may have difficulty capturing complex nonlinear relationships, while networks that are too deep tend to cause overfitting and waste of computational resources. $\varpi$ determines the number of temporal features that Gumbel-Softmax focuses on. A too small $\varpi$ makes the model tend to one-hot attention, causing insufficient feature attention, while too large $\varpi$ causes Gumbel-Softmax to converge to softmax, causing fuzzy attention that we wish to avoid from the beginning. Therefore, we choose the hyperparameter setting that is most favorable for performance, as shown in \autoref{setup}.

In summary, we propose CityEVCP, a deep learning model that incorporates dynamic and urban region influences, to achieve accurate citywide EV charging demand prediction. The experimental results empirically show that CityEVCP achieves state-of-the-art, with an average improvement of 10.47\% in three metrics and achieves the best fit in $\rm R^2$. Specifically, our method performs well on relatively long time prediction. The ablation experiments demonstrate the joint contribution of each module, and the full model achieves a 13.91\% improvement over the ablation models. We also explore the role of dynamic factors in improving the prediction performance of different city areas and illustrates the correct understanding of the proposed model. Furthermore, we experimentally explored the impact of hyperparameters.

\section{Conclusion}
\label{conclusion}
Accurate EV charging demand prediction is an important foundation for smart grids and transportation systems, helping to optimize the grid and guide the demand. How to incorporate multi-source heterogeneous external data into the model so that it can have a positive impact on improving the accuracy of EV charging demand prediction is a current research hotspot. To address the remaining research challenges, we propose an data-driven approach named CityEVCP that incorporates attentive hypergraph neural network for classified areas groups, adjacency graph attention, variable selection network and an improved Transformer encoder. CityEVCP (1) outperforms 10 representative methods by 10.47\% on average in three metrics and achieves the best fit in $\rm R^2$; (2) outperforms 7 ablation models comprehensively, demonstrating the enhancement brought by each introduced multi-source data or module. In our case study, we analyze the role of charging prices and temperatures for different areas of the city, validating the model's ability to correctly understand the dynamic auxiliary information. Moreover, we analyze the impact of hyperparameters and determined the preferred values in our case.

Furthermore, our research provides some perspectives on the management of smart energy and electric vehicle charging. First, achieving accurate charging demand prediction requires not only a focus on the charging data itself, but also on the external factors that influence EV charging choices. Therefore, information fusion from multiple sources at the government or manager level will be of significant help in collaborative and refined management.  Second, among the dynamic influences, price information contributes significantly to the accurate prediction of residential areas, while temperature information contributes to that of suburban and tourist areas. Third, associating charging facilities with similar properties in the surrounding helps in accurate node information propagation and collaborative area management.

Future research directions could be in the following aspects. (1) To explore more factors related to EV charging demand and their impacts, in order to capture the dynamics comprehensively and accurately. (2) To validate the proposed approach on other datasets including different EV usage scenarios, temperatures, prices, etc. (3) To build transferable networks that allow knowledge and prediction models to be transferred between regions or cities and make accurate demand prediction.

\printcredits

\bibliographystyle{unsrt}
\bibliography{ref}{}

\end{document}